\documentclass{article}
\usepackage[utf8]{inputenc}
\usepackage{amsmath,amssymb,amsthm} 

\usepackage[numbers]{natbib}

\makeatletter
\@ifundefined{date}{}{\date{}}

\theoremstyle{definition}
\newtheorem{example}{Example}

\DeclareMathOperator{\tr}{tr}
\newcommand{\ZZ}{\mathbb{Z}}
\newcommand{\KK}{\mathbb{K}}
\newcommand{\VV}{\mathcal{V}}
\newcommand{\xx}{\mathbf{x}}
\newcommand{\XX}{\mathbf{X}}
\newcommand{\bb}{\mathbf{b}}
\newcommand{\vv}{\mathbf{v}}
\newcommand{\ee}{\mathbf{e}}

\usepackage[scaled]{beramono}
\usepackage[T1]{fontenc}

\usepackage{listings}
\usepackage{xcolor}

\definecolor{codegreen}{rgb}{0,0.6,0}
\definecolor{codegray}{rgb}{0.5,0.5,0.5}
\definecolor{codepurple}{rgb}{0.58,0,0.82}
\definecolor{backcolour}{rgb}{0.95,0.95,0.92}

\lstdefinestyle{mystyle}{
    backgroundcolor=\color{backcolour},   
    commentstyle=\color{codegreen},
    keywordstyle=\color{magenta},
    numberstyle=\tiny\color{codegray},
    stringstyle=\color{codepurple},
    basicstyle=\ttfamily\footnotesize,
    breakatwhitespace=false,         
    breaklines=true,                 
    captionpos=b,                    
    keepspaces=true,                 
    numbers=left,                    
    numbersep=5pt,                  
    showspaces=false,                
    showstringspaces=false,
    showtabs=false,                  
    tabsize=2
}

\title{GAPS: Generator for Automatic Polynomial Solvers}
\author{
  Bo Li\\
  \texttt{\footnotesize prclibo@gmail.com}
  \and
  Viktor Larsson\\
  \texttt{\footnotesize viktor.larsson@inf.ethz.ch}
}
\makeatother

\begin{document}
\maketitle

\begin{abstract}
Minimal problems in computer vision raise the demand of generating efficient automatic solvers for polynomial equation systems. Given a polynomial system repeated with different coefficient instances, the traditional Gr\"obner basis or normal form based solution is very inefficient. Fortunately the Gr\"obner basis of a same polynomial system with different coefficients is found to share consistent inner structure. By precomputing such structures offline, Gr\"obner basis as well as the polynomial system solutions can be solved automatically and efficiently online. In the past decade, several tools have been released to generate automatic solvers for a general minimal problems. The most recent tool autogen from Larsson et al. is a representative of these tools with state-of-the-art performance in solver efficiency. GAPS wraps and improves autogen with more user-friendly interface, more functionality and better stability. We demonstrate in this report the main approach and enhancement features of GAPS. A short tutorial of the software is also included.
\end{abstract}

\section{Introduction}

GAPS\footnote{https://github.com/prclibo/gaps} wraps and improves autogen from Larsson et al. \cite{larsson2017efficient, larsson2018computational} which is first proposed in \cite{larsson2017efficient}. The software generates automatic polynomial solvers for a given multi-variate polynomials system with varying coefficients. It is originally intended to construct solvers for minimal problems in computer vision. Besides autogen, similar softwares also include Automatic Generator\footnote{https://github.com/PavelTrutman/Automatic-Generator} \cite{KBP08,trutman2015minimal} and PolyJam\footnote{https://github.com/laurentkneip/polyjam} \cite{kneip2015polyjam}. GAPS provides enhanced feature compared to these previous works in efficiency, usability and flexibility.

All of the above software share the same basic idea. For a given polynomial system, its Gr\"obner bases will in general have the same structure under different coefficient instances. The idea is now to first consider an instance of the problem where the coefficients are in some finite prime field $\ZZ_p$. This allows us to avoid any problems with numerical stability and we can easily compute the Gr\"obner basis, and use this to study the structure of the problem offline. This also allows us to recover which polynomial combinations of the input equations are necessary to find the Gr\"obner basis. Once this is known, we can leverage it to solve general instances with real coefficients. The assumption is now that the structure of the solution will be the same, but with different coefficients. The online phase of the solvers then only need to solve a linear system, essentially recovering the coefficients of the Gr\"obner basis, which allow us to turn the problem into an equivalent eigenvalue problem.


\section{Preliminaries}

We reuse the notation as in \cite{larsson2017efficient}. Define a polynomial
ring as $\KK[X]$ where $X=(x_{1,}\dots,x_{n})$ denotes variables
and $\KK$ denotes a field. Choices of $\KK$ that will
be used include $\mathbb{C}$ and $\ZZ_{p}$, with $p$ as
a prime number. The \emph{affine variety }of a set of polynomials
$F=\{f_{i}\}_{i=1}^{m}$ is denoted as
\begin{equation}
\VV(F)=\{\mathbf{x\in}\KK^{n}|f_{i}(x)=0,i=1,\dots,m\}.
\end{equation}

The sets of all polynomial combinations of the elements in F, i.e.
\begin{equation}
I(F)=\{p\in\KK[X]|p=\sum_{i}h_{i}f_{i},h_{i}\in\KK[X]\},\label{eq:ideal}
\end{equation}
forms an ideal in the polynomial ring $\KK[X]$. The Gr\"obner
basis is denoted as $G=\{g_{i}\}_{i=1}^{l}$ is a special sets of
generators of $I$ such that every leading term of $I$ is divisible
by the leading term of some member of $G$. Since a Gr\"obner basis
of an ideal $I$ exposes all the leading terms of $I$, it gives a
way of uniquely defining the remainders modulo $I$. This allows us
to define the normal form of a polynomial $p\in\KK[X]$ with
respect to $G$ as the unique remainder after division with $G$,
denoted $\overline{p}^{G}$. Note that $p\in I$ if and only if $\overline{p}^{G}=0$.
For a Gr\"obner basis $G$ the normal set is the set of all monomials
not divisible by any element in $G$. It is easy to see that the normal
form for any polynomial lies in the linear span of the normal set.

The quotient ring of $\KK[X]$ over $I$ is denoted as $\KK[X]/I$.
If an affine variety $\VV$ is zero dimensional (i.e. there
are finite solutions) then the corresponding quotient space $\KK[X]/I$
will be a finite dimensional vector space. The dimensionality of the
quotient ring is equal to the number of solutions.

\section{Approaches}
\subsection{The Main Idea}
The main idea of GAPS follows the conventional minimal solver generation
strategy similar to Automatic Generator \cite{KBP08}, of which the key step if to find the Gr\"obner basis and norm form of a polynomial system. Since such Gr\"obner basis computation is usually intractable for symbolic computation and numerically unstable for float-point complex number coefficients, it is more feasible to first look for Gr\"obner basis $G$ and norm form $B$ for a series of randomly generated coefficients from $\ZZ_p$. Based on previous work \cite{traverso1988grobner}, the union of $B$ of all $\ZZ_p$ instances forms the $B$ of the original polynomial system. Thus the overall pipeline of the solver can be summarized as:

\begin{enumerate}
\item Instantiate a set of polynomial coefficients from $\ZZ_p$.
\item Compute $B$ for each instances and collect the union.
\item Find all polynomials $\tilde F = \{m \cdot f | f \in F, m \in \mathsf{M}\}$ that will be linearly combined to construct action matrix and construct a template matrix whose columns corresponds to monomials and elements corresponds to coefficients. $\mathsf{M}$ is a set of monomials which are used to construct $\tilde F$.
\item Compute the RREF on the template matrix and the Gr\"obner basis are naturally obtained as rows of the RREF.
\item Construct action matrix to solve the unknown variable.
\end{enumerate}
The computation at step 2 can be done by mathematical software like Macaulay2. With $\mathsf{M}$ precomputed only once, $\tilde F$ can be constructed from $F$ directly for each minimal problem instances, which largely reduces the computation burden.

We next illustrate some details regarding instantiation on $\ZZ_p$ and the action matrix.

\subsection{Instantiation on $\ZZ_{p}$}

A well-known representative example of instantiation on $\ZZ_p$ is the conventional 5-pt problem as shown in Example~\ref{xmp:5-pt-1}.
\begin{example}[Instantiating a 5-point Problem]
5 point correspondences construct $5$ linear constraints on the Essential Matrix $E$. Denote the $E^{(i)}, i = 1,\dots, 4$ as the nullspace bases of the linear system such that $E = x E^{(1)} + y E^{(2)} + z E^{(3)} + E^{(4)}$. The polynomial constraints on $E$ are:
\begin{align}
\label{eq:det-e}
\det E &= 0,\\
\label{eq:eee}
2EE^\top E - \tr(EE^\top) E &= 0.
\end{align}
The coefficients of \eqref{eq:det-e}\eqref{eq:eee} can be instantiated by sampling $E^{(i)}$ from $\ZZ_p$. Since arbitrary $E^{(i)}$ correspond to a valid solution of \eqref{eq:det-e}\eqref{eq:eee}, each element of $E^{(i)}$ can be independently sampled from $\ZZ_p$. See \texttt{solvers/prob\_relpose\_5pt\_\_simple.m}.
\qed
\label{xmp:5-pt-1}
\end{example}
A drawback of such independent random sampling is that we do not know the value of $E$ corresponding to the instantiated coefficients. In addition, for a general minimal problem, an independent sampling does not guarantee that the inner structure and relationship of coefficients is preserved. This raises requirement for better sampling strategy in $\ZZ_p$. We next show how to instantiate in a generative strategy for 5-pt by the following series of examples
\begin{example}[Inverting a $\ZZ_p$ Element]
The inversion of a $\ZZ_p$ element $1/z$ can be obtained by the well-known Extended Euclidean algorithm, which computes in addition to the greatest common divisor of integers $z$ and $p$, also the coefficients of Bézout's identity, which are integers $x$ and $y$ such that
\begin{equation}
    z x + p y = \mathrm{gcd}(z, p) \equiv 1.
\end{equation}
The greatest common divisor is always $1$ due to $p$ being a prime number. Thus $x$ is the inversion of $z$ in $\ZZ_p$. See \texttt{generator/utils/zp\_inv.m}.
\qed
\label{xmp:inversion}
\end{example}

\begin{example}[Computing RREF and nullspace basis in $\ZZ_p$]
The computation of RREF of a matrix involves only operations of addition, subtraction, multiplication and division. Since division can be decomposed by multiplication and inversion as shown in Example~\ref{xmp:inversion}, RREF can be obtained in $\ZZ_p$ by replacing each operation to its $\ZZ_p$ version. See \texttt{generator/utils/zp\_rref.m}. Furthermore, the nullspace basis of a $\ZZ_p$ matrix can then be obtained from RREF.
\label{xmp:rref}
\qed
\end{example}

\begin{example}[Computing the Square Root in $\ZZ_p$]
The square root of a $\ZZ_p$ element $z$ can be computed by the Tonelli–Shanks algorithm. Note that $z$ has square root in $\ZZ_p$ if and only if the Euler's criterion hold as:
\begin{equation}
    z^\frac{p - 1}{2} \equiv 1 \mod p.
    \label{eq:euler-criterion}
\end{equation}
See \texttt{generator/utils/zp\_sqrt.m}.
\qed
\end{example}

\begin{example}[Instantiating a $\ZZ_p$ Unit Vector]
Given a random $\ZZ_p$ vector $z$, we first compute its squared norm $\|z\|^2$ in $\ZZ_p$. If $\|z\|^2$ passes Euler's criterion \eqref{eq:euler-criterion}, it is not difficult to normalize $z$ in $\ZZ_p$ by computing $z \cdot \frac{1}{\sqrt{\|z\|^2}}$ in $\ZZ_p$. See \texttt{generator/utils/zp\_rand\_unit.m}.
\label{xmp:unit-vector}
\qed
\end{example}

\begin{example}[Instantiating a Rotation Matrix]
Following Example~\ref{xmp:unit-vector} we first instantiate a unit quaternion $[\sigma\ u]$ in $\ZZ_p$. A rotation matrix in $\ZZ_p$ is then obtained by consider the Rodrigues' formula in $\ZZ_p$:
\begin{equation}
    R = 2 (u u^\top - \sigma [u]_\times) + (\sigma^2 - \|u\|^2) I.
    \label{eq:rodrigues}
\end{equation}
\qed
\end{example}

\begin{example}[A Generative 5-Point Problem Instance]
Consider a camera pair with $P_1 = [I\ 0_{3\times1}]$ and $P_2 = [R\ t]$. It is not difficult to instance $P_2$ in $\ZZ_p$ following the previous examples. A 3D point instance $X$ is projected into the two cameras as $x_1$ and $x_2$. The projection involves operation addition, multiplication and division in $\ZZ_p$. Division in $\ZZ_p$ can be executed by multiplication and inversion. In this way we can obtained $5$ pairs of correspondences to construct a 5-point problem instance in $\ZZ_p$. The corresponding groundtruth Essential Matrix is also known as $E = [t]_\times R$. $E^{(i)}$ can also be deduced by computing RREF as in Example~\ref{xmp:rref}.
\qed
\end{example}

The above examples covers most of the operation for users to generate other minimal problem instances in $\ZZ_p$.

\subsection{The Action Matrix Method}

The action matrix method has been one of the most widely used methods to solve multivariate polynomial system in minimal problems. It reduce a minimal problem to an eigenvalue problem for which there exist good numerical methods.

Consider the operator $T_\alpha : \KK[X]/I \rightarrow{} \KK[X]/I$:
\begin{equation}
    T_\alpha[p(\xx)] = \overline{\alpha(\xx) p(\xx)}^G,
\end{equation}
with $p(\xx), \alpha(\xx)\in \KK[X]$. Usually $\alpha(\xx)$ is selected as a monomial on $\xx$. $\KK[X]/I$ is a linear space and the normal form set $B$ forms bases of the space. If we concatenate bases in $B$ as a vector $\bb(\xx) = [b_1(\xx), \dots, b_m(\xx)]^\top$, any element $p(\xx)$ in $\KK[X]/I$ can be denoted as a vector $\vv$ by:
\begin{equation}
    p(\xx) = \vv^\top \bb(\xx).
\end{equation}
Denote $\ee_i = [0, \dots, 1, \dots, 0]$ as the standard vector basis and consider the $T_\alpha[b_i(\xx)]$:
\begin{equation}
    T_\alpha[b_i(\xx)] = \overline{\alpha(\xx)b_i(\xx)}^G = \sum_j m_{i, j} b_j(\xx)
\end{equation}
The operation $T_\alpha[p(\xx)]$ can be then expressed as a linear transform:
\begin{multline}
    T_\alpha[\vv] = T_\alpha[p(\xx)] = \overline{\alpha(\xx) \vv^\top \bb(\xx)}^G = \overline{\vv^\top (\alpha(\xx) \bb(\xx))}^G = \vv^\top \overline{\alpha(\xx) \bb(\xx)}^G\\
    = \vv^\top \begin{bmatrix}
    \sum_j m_{1, j} b_j(\xx)\\
    \sum_j m_{2, j} b_j(\xx)\\
    \cdots
    \end{bmatrix} = \vv^\top M \bb(\xx)
\end{multline}
For any $\xx \in \VV$ we have $f(\xx) = \overline{f(\xx)}^G$. Thus for any solution $\xx \in \VV$, we can remove the remainder operator and obtain:
\begin{align}
    &\alpha(\xx) \vv^\top \bb(\xx) = \overline{\alpha(\xx) \vv^\top \bb(\xx)}^G = \vv^\top M \bb(\xx), \quad \forall \vv;\\
    \Rightarrow \quad &\alpha(\xx) \bb(\xx) = M \bb(\xx).
\end{align}
The eigenvalues and eigenvectors of $M$ is the value of $\alpha(\xx)$ and $\bb(\xx)$ at each of the solution $\xx \in \VV$. If we can construct the action matrix $M$ we convert the problem to an eigenvalue problem, and hence we have reduced the solving of the system of polynomial equations to a linear algebra problem.

If $\alpha b_i \in B$, denoted as $b_k$ it is easy to construct row $m_{i, :} = e_k$ where $e_k$ is the $k$-th standard vector basis. Otherwise, we call $r_i = \alpha b_i$ a \textit{reducible} monomial. We have:
\begin{equation}
    r_i - \overline{r_i}^G = r_i - \sum_j m_{i, j} b_j \in I.
    \label{eq:red-quot}
\end{equation}
Thus there exist some polynomials $h_{i, j} \in \KK[X]$ such that
\begin{equation}
    r_i - \sum_j m_{i, j} b_j = \sum_j h_{i, j} f_j.
\end{equation}
\eqref{eq:red-quot} is a linear combination of some $h_{i, j} f_j$. The coefficients of $h_{i, j}$ depend on specific problem instances. We next show a universe solution to construct $r_i - \sum_j m_{i, j} b_j$. Decompose $h_{i, j}$ to a set of monomials $\mathsf{m}(h_{i, j})$, without coefficients. Denote $\mathsf{M}_i = \bigcup_{j} \mathsf{m}(h_{i, j})$ and
\begin{equation}
    \tilde F = \bigcup_i \{ m \cdot f_i | m \in \mathsf{M}_i \}.
\end{equation}
\eqref{eq:red-quot} is a linear combination of elements in $\tilde F$. Stack all polynomials in $\tilde F$ as a linear system $C \XX = 0$ where $\XX$ is a vector of all the monomials occurring in $\tilde F$ and $C$ contains the corresponding coefficients. $C$ is referred as the (elimination) template matrix. We reorder the system as follows:
\begin{equation}
    C \XX = \begin{bmatrix}C_E & C_R & C_B\end{bmatrix} \begin{bmatrix}
    \xx_E\\ \xx_R\\ \xx_B
    \end{bmatrix} = 0,
\end{equation}
where monomials are grouped into excessive monomials $\xx_E$, reducible monomials $\xx_R$ and basis monomials $\xx_B$. \eqref{eq:red-quot} can then be stacked as:
\begin{equation}
    \begin{bmatrix}0 & I & -M'\end{bmatrix} \begin{bmatrix}
    \xx_E\\ \xx_R\\ \xx_B
    \end{bmatrix} = 0.
\end{equation}
where $M'$ is stacked by $m_{i, :}$ for $b_i \notin B$. Hence by computing the RREF of $C$ and extracting the bottom rows, we can obtain $M'$ and then $M$.

The remaining problem is the computation of $\mathsf{M}_i$. Since the set of $\mathsf{M}_i$ remain the same regardless of the problem instances, it is convenient to precompute $\mathsf{M}_i$ given a minimal problem to reduce computational burden.

Following autogen, $\mathsf{M}_i$ is computed from a problem instance in $\ZZ_p$. $h_{i, j}$ for this instance can be obtained using softwares like \texttt{Macaulay2}. $\mathsf{M}_i$ is then extracted. To avoid missing monomials with zero coefficients, this procedure is repeated multiple times and the union of $\mathsf{M}_i$ is taken.

\section{Enhancement}

\subsection{Syzygy Reduction}

The RREF computation on a template matrix is one of the sources of computational burden in a minimal solver. Therefore, several previous works have been proposed to reduce the size of template matrix for efficiency. 

\cite{larsson2017efficient} proposed to reduce the degree of $\mathbf{h}_i = [h_{i, j}]_j$ to reduce the size of $\mathsf{M}_i$. Define the \textit{first syzygy module} \cite{CLS} of $F$ as:
\begin{equation}
    \mathcal{M} = \{ \mathbf{s} \in \ZZ_p[X] | \sum_j s_j f_j = 0, f_j \in F \}.
\end{equation}
For a $\ZZ_p$ instance, the Gr\"obner basis of $\mathcal{M}$ is computed as $G_\mathcal{M}$. \cite{larsson2017efficient} proposes that $\overline{\mathbf{h}_i}^{G_\mathcal{M}}$ is a simpler replacement for $\mathbf{h}_i$.

\subsection{Upper Triangular Matrix Reduction}

Another strategy to reduce the size of template matrix is to find an upper triangular structure in $C_E$. If we can reorder some columns in $C_E$ and some rows in $C$ such that the upper-left submatrix of $C$ is upper triangular, i.e. $\begin{bmatrix} U & V \\ W & X \end{bmatrix}$, the template matrix can be efficiently reduced to $\begin{bmatrix} U & V \\ 0 & C' \end{bmatrix}$, with $C' = X - W U^{-1} V$ as the new template matrix.

Note that we only reorder columns in $C_E$ to keep the reducibles and basis. Although reordering a matrix to be upper triangular is known to be NP-complete \cite{fertin2015obtaining},
it is not difficult to execute a greedy strategy to find out a non-optimal upper triangle.

\subsection{Characteristic Polynomial}

For large action matrix, the computation of eigenvalues and eigenvectors is another source of computation burden in a minimal solver. A enhancement for the efficiency is to convert the action matrix to its characteristic polynomial which can be solved by Sturm sequence very fast. This problem is recently summarized in \cite{bujnak2012making} and the approach proposed by \cite{danilevskiy1961numerical} is used in GAPS to generate characteristic polynomial.

Note that it is admitted widely that solutions from characteristic polynomials is numerically less stable than those from action matrix.

\subsection{Coefficient Simplification}
The expansion of coefficients in $F$ might be complicated and contains many repeated computation. Typical optimization of computation include the Horners' algorithm and its multivariate version. By wrapping autogen, GAPS, uses the Matlab shipped $O(n)$ optimization strategy to simplify the coefficients computation.

\subsection{Symmetries}
The generator can automatically identify and use (partial) p-fold symmetries in the equation system as described in \cite{ask2012exploiting,kuang2014partial,larsson2016uncovering}. Currently only variable-aligned symmetries (e.g.~sign changes) are automatically detected. Note that if symmetries are detected, the generated solvers only returns one set of solutions, and the user will manually have to compute the symmetric solutions by applying the symmetry (so e.g.~by flipping signs). Common occurrences of these symmetries in computer vision are due to the sign invariance of the quaternion parameterization or the ambiguity between the sign of the focal length and the camera orientation.

\section{A Short Tutorial}

To construct a polynomial solver, inherit the problem (see \texttt{generator/problem.m}) class to specify your polynomial system. Implement three functions in your inheritance.

\texttt{[in, out] = gen\_arg\_subs(obj)} creates two structs corresponding to input/output variables. Field names of the struct will be argument names used in the generated function. Field values are sym variables that will be used to denote polynomials.

\texttt{[eqs, abbr\_subs] = gen\_eqs\_sym(obj)} creates sym equation polynomials.

\texttt{[in\_zp, out\_zp] = rand\_arg\_zp(obj, p)} generates random sample on Zp for variables in this problem. Field names in \texttt{in\_zp} and \texttt{out\_zp} correspond to the known and unknown symbolic variables in the polynomials.

\texttt{[in\_rl, out\_rl] = rand\_arg\_rl(obj)} generates random sample on real field for variables in this problem. You should instantiate this member function for your problem. Field names in \texttt{in\_rl} and \texttt{out\_rl} correspond to the known and unknown sym variables in the polynomials.

\subsection{A Minimal Example}

We use GAPS to solve the 5-point problem described in Example~\ref{xmp:5-pt-1}.
\begin{example}

Below is a simple example on solving the 5-pt problem:
\begin{lstlisting}[language=Matlab]
classdef prob_pc_relpose_5p_nulle_ne__simple < problem
    % A simple instance to construct a pinhole camera 5-pt relative pose
    % estimation problem solver.
    methods
        function [in, out] = gen_arg_subs(obj)
            % Each field in `in_subs/out_subs` will become an input/output
            % argument in the generated solver.
            %
            % `in_subs.NE` is a 4x3x3 matrix made up of NEijk symbols. The
            % created solver will expect input argument NE to be 4x3x3
            % matrix and fill its element values to NEijk respectively.
            %
            % See your favorite 5-pt paper for details.
            %
            % Base vectors of the null space of the essential matrix.
            in.NE = sym('NE%d%d%d', [4, 3, 3]);
            % The weights
            out.w = sym('w%d', [3, 1]);
        end
        function [eqs_sym, abbr_subs] = gen_eqs_sym(obj)
            [in, out] = gen_arg_subs(obj);
            NE = permute(in.NE, [2, 3, 1]);
            NE = reshape(NE, 9, 4);
            E = reshape(NE * [out.w; 1], 3, 3);
            eqs_sym = sym([]);
            
            % Construct polynomial system as symbolics
            eqs_sym(1) = det(E);
            Et = transpose(E);
            te = 2*(E*Et)*E - trace(E*Et)*E;
            eqs_sym(2:10) = te(:);
            
            % abbr_subs (Abbreviation substitution) is used to declare
            % intermediate variables when computing coefficients and their
            % expansions. This this very simple case we will not use it.
            abbr_subs = struct([]);
        end
        function [in_zp, out_zp] = rand_arg_zp(obj, p)
            % For 5-pt problem, arbitrary value of NE always corresponds to
            % valid polynomial system. Therefore we just instantiate a Zp
            % case by random integer. However, usually this does not hold
            % for a general minimal problem. You need to consider the
            % geometry constraint during instantiation.
            %
            % in_zp/out_zp is expected to have same fields as of in_subs
            % and out_subs returned by `gen_arg_subs`. But the field values
            % are symbolic integers.
            in_zp.NE = sym(randi([1, p - 1], [4, 3, 3]));
            % out_zp can be omitted as it is not used right now.
            out_zp = struct();
        end
        function [in_rl, out_rl] = rand_arg_rl(obj)
            error(['This is similar to `rand_arg_zp` with real field values. ',...
                'We are not using it here as it is for benchmarking']);
        end
    end
end
\end{lstlisting}
To solve this problem, use scripts like the following:
\begin{lstlisting}[language=Matlab]
opt = default_options();
opt.M2_path = '/Users/li/workspace/Macaulay2-1.13/bin/M2';
opt.optimize_coefficients = true;

prob_fn = @prob_pc_relpose_5p_nulle_ne__simple;
[solv, opt] = generate_solver(prob_fn, opt);
\end{lstlisting}
Sample files can be found in folder \texttt{problems/}.
\qed
\end{example}

\subsection{A Full Example}
We use GAPS to solve the 4-point relative pose problem with a rotation angle \cite{MartLi19} as a full example.

\begin{example}
Parameterize rotation $R$ with $[\sigma\ u]^\top$ by \eqref{eq:rodrigues}. The knowns include $4$ point correspondences $q_i', q_i''$ and $\sigma$ and the unknowns are vector $u$. For any triplet of point correspondences $i, j, k$, construct
\begin{equation}
    F_{ijk} = \begin{bmatrix}
    q_j''^\top R p_{ij}' & p_{ij}''^\top R q_j'\\
    q_k''^\top R p_{ik}' & p_{ik}''^\top R q_k'
    \end{bmatrix},
\end{equation}
where $p_{ij}' = q_i' \times q_j'$ and similarly for $p_{ij}''$. The problem polynomial is written as
\begin{equation}
    F = \{\det F_{123}, \det F_{234}, \det F_{341}, \det F_{412}, u^\top u + \sigma^2 - 1\}.
\end{equation}

We define a Matlab class corresponding to this problem by inheriting from the \texttt{problem} class and declare the input and output arguments. In GAPS, we use the terms \textit{arguments} (arg) to refer semantic arguments like a scalar $s$, a vector \texttt{u} or a matrix \texttt{q}. \texttt{in\_subs} and \texttt{out\_subs} will correspondingly contain a field with the same name as the argument. The value of the field is Matlab symbols representing each scalar value in the argument, which is referred as \textit{variables} (var).

\begin{lstlisting}[language=Matlab]
classdef prob_pc_relpose_4pra_sir2__example < problem
    methods
        function [in_subs, out_subs] = gen_arg_subs(obj)
            in_subs.q = sym('q%d%d', [3, 4]);
            in_subs.qq = sym('qq%d%d', [3, 4]);
            in_subs.s = sym('s');
            out_subs.u = sym('u%d', [3, 1]);
        end
        function [eqs_sym, abbr_subs] = gen_eqs_sym(obj)
            [in, out] = obj.gen_arg_subs();
            u = out.u;
            s = in.s;
            R = 2 * (u * u.' - s * skew(u)) + (s * s - u.' * u) * eye(3);
            
            eqs_sym = sym([]);
            abbr_subs = struct();
            for i = 1:obj.N
                j = mod(i, obj.N) + 1;
                k = mod(i + 1, obj.N) + 1;
                
                [F, F_subs] = obj.Fijk(i, j, k, in.q, in.qq, R);
                eqs_sym(end + 1) = det(F);
                 % CATSTRUCT is a utility function shipped in GAPS
                abbr_subs = catstruct(abbr_subs, F_subs);
            end
            eqs_sym(end + 1) = out.u.' * out.u + in.s .* in.s - 1;
        end
        % To be continued
\end{lstlisting}
\texttt{abbr\_subs} is used to user-define substitution of repeated computation patterns such as $p_{ij}'$. \texttt{abbr\_subs} is a structure with similar design with \texttt{in\_subs} and \texttt{out\_subs} except that the field values are not symbols but symbol expressions. When evaluating coefficients, \texttt{abbr\_subs} will be evaluated first and substitute into coefficients, avoiding duplicated computation defined as symbol expressions in the fields. See how it is used in the function \texttt{obj.Fijk}.
\begin{lstlisting}[language=Matlab]
        % Cont'd
        function [F, abbr_subs] = Fijk(obj, i, j, k, q, qq, R)
            pij = sym(sprintf('p_%d_%d_%%d', i, j), [3, 1]);
            pik = sym(sprintf('p_%d_%d_%%d', i, k), [3, 1]);
            ppij = sym(sprintf('pp_%d_%d_%%d', i, j), [3, 1]);
            ppik = sym(sprintf('pp_%d_%d_%%d', i, k), [3, 1]);
            
            F = [qq(:, j).' * R * pij, ppij.' * R * q(:, j);
                qq(:, k).' * R * pik, ppik.' * R * q(:, k)];
            pij_val = skew(q(:, i)) * q(:, j);
            pik_val = skew(q(:, i)) * q(:, k);
            ppij_val = skew(qq(:, i)) * qq(:, j);
            ppik_val = skew(qq(:, i)) * qq(:, k);
            
            c = num2cell([pij_val; pik_val; ppij_val; ppik_val]);
            f = cellstr(string([pij, pik, ppij, ppik]));
            abbr_subs = cell2struct(c, f);
        end
        % To be continued
\end{lstlisting}
We next show how to generate $\ZZ_p$ instances for this problem. \texttt{in\_zp} and \texttt{out\_zp} have the same design with \texttt{in\_subs} and \texttt{out\_subs} except that the field value are Matlab symbolic integers.
\begin{lstlisting}[language=Matlab]
        % Cont'd
        function [in_zp, out_zp] = rand_arg_zp(obj, p)
            quat = zp_rand_unit(4, p);
            s = quat(1);
            u1 = quat(2);
            u2 = quat(3);
            u3 = quat(4);
            u = [u1; u2; u3];
            
            % Make use of some utility functions.
            R = zp_quat2dcm(quat, p);            
            t = zp_rand_unit(3, p);
            
            Q = sym(randi(p, [3, obj.N]));
            % Yes in Matlab 2018 you don't need to REPMAT.
            QQ = mod(R * Q + repmat(t, [1, obj.N]), p);
            
            % We ignore to make Q and QQ homogenous
            in_zp.q = Q; in_zp.qq = QQ;
            in_zp.s = s; out_zp.u = u;
        end
        % To be continued
\end{lstlisting}
If we would like to call \texttt{benchmark\_solver} to benchmark or test the solver, it is also useful to implement \texttt{obj.gen\_arg\_rl} to sample a problem instance in $\mathbb{R}$. \texttt{in\_rl} and \texttt{out\_rl} have the same design with \texttt{in\_subs} and \texttt{out\_subs} except that the field value are Matlab numeric double.
\begin{lstlisting}[language=Matlab]
        function [in_rl, out_rl] = rand_arg_rl(obj)
            quat = normc(rand([4, 1]));
            s = quat(1);
            u1 = quat(2);
            u2 = quat(3);
            u3 = quat(4);
            u = [u1; u2; u3];
            
            R = 2 * (u * u.' - s * skew(u)) + (s * s - u.' * u) * eye(3);
            t = normc(rand([3, 1]));
            Q = rand([3, obj.N]);
            QQ = R * Q + repmat(t, [1, obj.N]);
            in_rl.q = Q; in_rl.qq = QQ;
            in_rl.s = s; out_rl.u = u;
        end
    end % methods
end % classdef
\end{lstlisting}
Sample files can be found in folder \texttt{problems/}.
\qed
\end{example}

\bibliographystyle{plainnat}
\bibliography{main}

\end{document}